%% file: preprint.tex
\documentclass[runningheads]{llncs}
\usepackage{graphicx}
\usepackage{enumitem}
\usepackage{hyperref}
\usepackage{cleveref}
\usepackage{paralist}
\usepackage{tabularx}
\usepackage{booktabs}
\usepackage{xcolor}

\usepackage[misc]{ifsym}
\Crefformat{figure}{#2Fig.~#1#3}
\Crefmultiformat{figure}{Figs.~#2#1#3}{ and~#2#1#3}{, #2#1#3}{ and~#2#1#3}

\hypersetup{pdfcreator={Me}}

\title{DinoBloom: A Foundation Model for Generalizable Cell Embeddings in Hematology}
\titlerunning{DinoBloom}

\author{Valentin Koch\textsuperscript{1,2,$\star$}, Sophia J. Wagner\textsuperscript{1,2,$\star$}, Salome Kazeminia\textsuperscript{1,2}, Ece Sancar\textsuperscript{1,2}, Matthias Hehr\textsuperscript{2,3}, Julia A. Schnabel\textsuperscript{1,2,4}, Tingying Peng\textsuperscript{1,2}, Carsten Marr\textsuperscript{2,\Letter}}

\authorrunning{V. Koch et al.}

\institute{School of Computation, Information and Technology, Technical University of Munich, Munich, Germany
\and Computational Health Center, Helmholtz Munich, Munich, Germany \and
Dr. von Haunersches Kinderspital, Ludwig-Maximilians-University Munich, Munich, Germany \and
School of Biomedical Engineering and Imaging Sciences, King’s College London, London, UK
}

\begin{document}
\maketitle
\let\thefootnote\relax\footnotetext{\textsuperscript{$\star$} Contributed equally}
\let\thefootnote\relax\footnotetext{\textsuperscript{\Letter}\href{mailto:carsten.marr@helmholtz-munich.de} {carsten.marr@helmholtz-munich.de}}
%
%

%
\begin{abstract}
In hematology, computational models offer significant potential to improve diagnostic accuracy, streamline workflows, and reduce the tedious work of analyzing single cells in peripheral blood or bone marrow smears. However, clinical adoption of computational models has been hampered by the lack of generalization due to large batch effects, small dataset sizes, and poor performance in transfer learning from natural images. To address these challenges, we introduce DinoBloom, the first foundation model for single cell images in hematology, utilizing a tailored DINOv2 pipeline. Our model is built upon an extensive collection of 13 diverse, 
publicly available datasets of peripheral blood and bone marrow smears, the most substantial open-source cohort in hematology so far, comprising  over 380,000 white blood cell images.
To assess its generalization capability, we evaluate it on an external dataset with a challenging domain shift. We show that our model outperforms existing medical and non-medical vision models in (i) linear probing and \textit{k}-nearest neighbor evaluations for cell-type classification on blood and bone marrow smears and (ii) weakly supervised multiple instance learning for acute myeloid leukemia subtyping by a large margin. 
A family of four DinoBloom models (small, base, large, and giant) can be adapted for a wide range of downstream applications, be a strong baseline for classification problems, and facilitate the assessment of batch effects in new datasets. All models are available at 
\href{https://github.com/marrlab/DinoBloom}{github.com/marrlab/DinoBloom}.

\end{abstract}

\input{preprint/introduction}

\input{preprint/datasets}
\input{preprint/methods}

\input{preprint/results}
\input{preprint/conclusion}
\bibliographystyle{splncs04}
\bibliography{references}
\end{document}

%% file: preprint/introduction.tex
\section{Introduction}

Hematology, the study of blood and blood-related diseases, relies heavily on the microscopic examination of peripheral blood and bone marrow smears. This practice is integral to diagnosing hematological diseases, such as acute myeloid leukemia (AML)~\cite{khoury20225th,arber2022international}. 
Currently, differential blood counts still rely on manual cytomorphological analysis of at least 200 individual white blood cells (WBC) per patient, where 
%
%
exact evaluation is crucial for early and precise diagnosis~\cite{houwen2001differential}. This labor-intensive process has resisted automation, remaining a domain for trained experts. 
However, it suffers from significant intra- and inter-expert variability, complicating diagnosis in environments that lack trained personnel~\cite{font2015interobserver}. 
%

Recent advances in deep learning propose solutions to the challenges in hematology, such as classifying leukemia subtypes from microscopic images~\cite{matek2019human,matek2021highly}.
However, the transition from manual to automated analysis requires robust models that can deal with limited data, strong batch effects, and largely varying cell phenotypes. In particular, for weakly-supervised settings, such as multiple instance learning (MIL) for patient-level disease prediction, a strong feature extractor for single blood cells is necessary as supervised learning is not possible. 
%
%

Still, most approaches so far rely on supervised training sets with corresponding datasets.
Rastogi et al.~\cite{LeuFeatx} train a convolutional neural network (CNN) as a feature extractor on the AML Matek dataset \cite{matek2019human} on 18,365 single-cell images from peripheral blood smears. 
Hehr et al.~\cite{hehr2023explainable} train a feature extractor fully supervised on a cell classification task on additional data from the same domain and subsequently train the MIL aggregation model.
%
%

Large-scale self-supervised training on diverse datasets has transformed the domain of computer vision on natural images~\cite{he2020momentum,caron2021emerging}. In the medical imaging domain, especially, in histopatholoy, domain specific self-supervised representation learning on large sets of unlabeled images~\cite{WANG2023102645,ctranspath} has shown to improve downstream tasks in MIL settings~\cite{wagner2023transformer}. To this end, DINO~\cite{caron2021emerging} and its successor DINOv2~\cite{oquab2023dinov2} have emerged as a pipeline well-suited to train these feature extractors~\cite{chen2023generalpurpose,vorontsov2024virchow,phikon}. However, to date, there is no comparable effort tackling the challenges in the domain of hematology.

We propose DinoBloom (Dino Blood Model), a model family based on
vision transformers trained with a customized DINOv2~\cite{oquab2023dinov2} pipeline to provide rich visual features for hematological single-cell image analysis. 
The models are able to extract highly predictive features even on unseen datasets that can be used for few-shot classification, multiple instance learning or cell embeddings potentially characterizing disease profiles while offering explainable features for enhanced interpretability. The main contributions of our work are:
\begin{compactitem}

    \item We introduce DinoBloom, the first large-scale self-supervised trained models designed explicitly for single-cell hematology image analysis.
    \item We assemble the largest cohort in hematology comprising $13$ datasets of peripheral blood and bone marrow smears.
    \item We show that DinoBloom models are effective in capturing diverse visual features of single cells across tasks on both in-domain and out-of-domain datasets for cell-type classification and leukemia subtyping.
    \item We provide open access to all DinoBloom models as well as the source code and parameters used for training, encouraging the research community to collaboratively build upon our work.
\end{compactitem}

\begin{figure}[t]
\includegraphics[width=\textwidth]{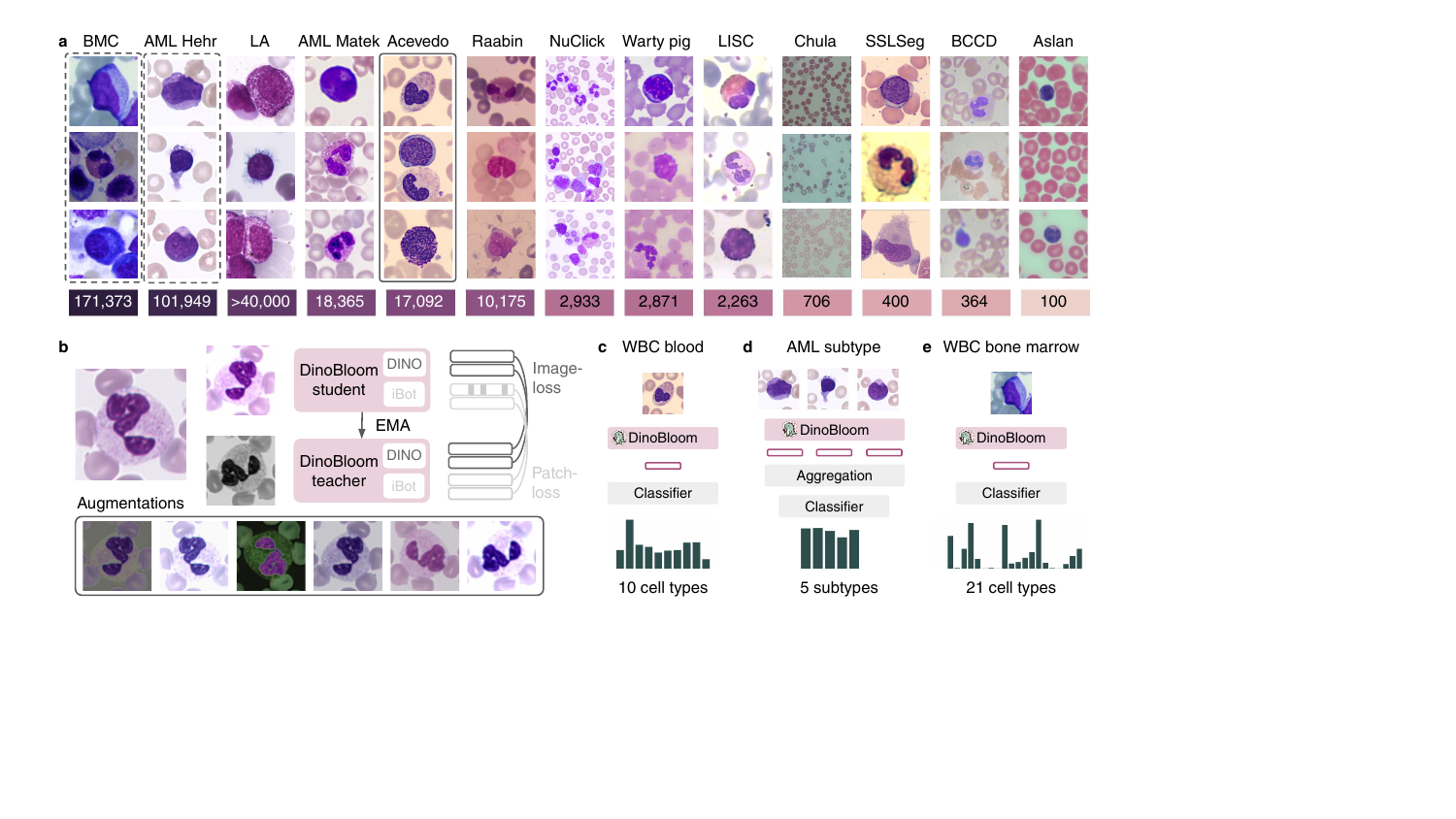}
\caption{Data and model overview of our pipeline. (a) All $13$ datasets used in this study: dashed lines indicate datasets split into training data for DinoBloom and test data for downstream evaluations, continuous line indicates the dataset was completely held out for testing purposes. (b) Modified DINOv2 pipeline without local crops for model training. We evaluate the performance on three downstream tasks: (c) WBC type classification on the external dataset Acevedo, (d) AML subtype classification via multiple instance learning, and (e) bone marrow WBC type classification. 
} \label{fig:overview}
\end{figure}

%% file: preprint/datasets.tex
\section{Datasets}

To the best of our knowledge, we collected the largest hematology cohort with $13$ publicly available datasets of in total 
380,000 blood cell images, fusing the domains of peripheral blood and bone marrow smears (\Cref{fig:overview}a). It consists of the following datasets: 
The \textit{Bone Marrow Cytomorphology} (BMC) dataset contains 171,373 de-identified, expert-annotated cells from bone marrow smears of $945$ patients~\cite{matek2021highly}. In contrast to all other datasets included in this study, the bone marrow smears were stained using the May-Grünwald-Giemsa/Pappenheim stain.
The \textit{AML Hehr} dataset includes 81,214 single-cell images from $189$ patients, covering four genetic AML subtypes and a healthy control group, sourced from annotated patient-level blood smears.
    The \textit{LA} (LabAnonymous) dataset consists of over 40,000 images of peripheral blood smears.
    The dataset will be made publicly available through a different publication. 
    The \textit{AML Matek} dataset consists of 18,365 expert-labeled single-cell images with $15$ heavily imbalanced classes taken from peripheral blood smears of $100$ patients diagnosed with AML, as well as 100 healthy patients~\cite{matek2019single,matek2019human}. 
    %
    %
    %
    \textit{Acevedo} encompasses 17,092 images of WBCs labeled into 11 classes~\cite{acevedo2020dataset}.
    The \textit{Raabin} dataset features 10,175 cropped WBCs, which are labeled by experts into five classes~\cite{kouzehkanan2021raabin}. 
    The \textit{NuClick} dataset was created from 11,000 WBC images of four classes 
    to generate 2,689 images of artificially overlapping nuclei~\cite{koohbanani2020nuclick}.
    The \textit{Warty pig} dataset contains 1,408 cropped and classified, plus 1,463 augmented WBC images of juvenile Visayan warty pigs~\cite{wartypig}.
    The public part of the \textit{LISC} dataset consists of $157$ WBC images as well as several augmented versions of them, totaling $2263$ images~\cite{rezatofighi2011automatic}. 
    \textit{Chula} \cite{naruenatthanaset2023red} is a red blood cell segmentation dataset and holds 706 single images.
    The \textit{SSLSeg} datasets contain $300$ images of WBCs, stemming from two different sources (200/100 images)~\cite{zheng2018fast}.
    The \textit{Blood Cell Count and Detection} (BCCD) dataset was created to detect blood cells and includes $364$ images with bounding box labels \cite{mohamed2012efficient}.
    The \textit{Aslan} blood cell detection dataset offers $100$ images of white and red blood cells taken from a light microscope \cite{aslan_blood_cell_detection}. 

%% file: preprint/methods.tex
\section{Methods}

\textbf{DINOv2 finetuning.}
We train our models that are based on vision transformers (ViT)~\cite{vit} in different sizes using the DINOv2 framework. Following~\cite{roth2024low}, we use the pretrained checkpoints to efficiently finetune the vision transformer on our multi-cohort dataset. 
The self-supervised learning framework DINOv2 employs a teacher-student architecture with an image-based loss on the class token of the DINO head and a patch-based loss on the class token of the masked patches from the iBot head (\Cref{fig:overview}b). 
We remove the global-local crop loss as we found it hampers performance when learning representations on the single-cell images in blood and bone marrow datasets. 
Images are resized to 224$\times$224 pixels. 
The models are trained on $8$ NVIDIA A100-SXM4-40GB GPUs with an AMD EPYC 7742 64-Core CPU. 
All models show similar convergence patterns and reached their peak performance between 4,000 and 8,000 iterations, after which downstream task performance drops slightly. 
Depending on the model and corresponding batch size, 4,000 iterations cover the training set between $1.7$ (batch size 208, DinoBloom-G) up to $10$ times (batch size 1,216, DinoBloom-S). 
Used batch size, feature dimension, and training time, as well as the number of parameters, can be inferred from~\Cref{tab:model_batch_sizes}.

\begin{table}[t]
    \centering
    \caption{Training configuration of DinoBloom models.}
    \begin{tabular}{l|c|c|c|rl}
        \hline
        Model & Batch size & Train time& Feature dim & \multicolumn{2}{c}{\#params}  \\ 
        \hline
        DinoBloom-S & $1216$ & 1:30 h & $384$ & $22$ & M \\ 
        DinoBloom-B & $960$ & 0:45 h & $768$ & $86$ &M\\
        DinoBloom-L & $448$ & 1:00 h& 1024 & $304$& M\\
        DinoBloom-G & $208$ & 4:00 h & 1536 & 1136 &M\\
        \hline
    \end{tabular}
    \label{tab:model_batch_sizes}
\end{table}

\textbf{Train and test data.}
We train the DinoBloom models on all datasets except the Acevedo dataset, which is kept as external test set. The other two datasets used for evaluation were split into train/test $(80/20)$. Only training data was used to train our DinoBloom models. We evaluate the downstream task performance using the same train/test split. The following datasets and settings are used:
\begin{compactitem}
    \item Acevedo is the smallest of the peripheral blood datasets with complete image-level classification exhibiting a strong batch effect (\Cref{fig:overview}a), hence serving as a good measure of the generalization capabilities of our model. 
    \item The AML Hehr dataset is divided into train/test (80/20) on patient level. It includes 101,949 WBCs from 242 patients across four AML subtypes: CBFB::MYH11, NPM1, PML::RARA, and RUNX1::RUNX1T1 and a healthy control class. 
    \item The BMC dataset is split into train/test $(80/20)$ as it is the only bone marrow dataset. 
    It contains $21$ heavily imbalanced classes (\Cref{fig:overview}e). 
\end{compactitem}

\begin{table}[t]
    \centering
    \caption{Evaluation on peripheral blood: Image-level WBC classification on Acevedo dataset and patient-level AML subtyping on AML Hehr dataset.  Best results are marked in bold, second best results are underlined. Standard deviation in the ABMIL setting was obtained by 5-fold cross-validation. Performance is measured in weighted F1-score (wF1) and balanced Accuracy (bAcc).}
    \label{tab:hema}
    \begin{tabular}{l|cc|cc|cc|cc}
        \hline
        & \multicolumn{6}{c|}{Acevedo} & \multicolumn{2}{c}{AML Hehr} \\ \hline
        & \multicolumn{2}{c|}{1-NN} 
        & \multicolumn{2}{c|}{20-NN} 
        & \multicolumn{2}{c|}{Linear probe} 
        & \multicolumn{2}{c}{ABMIL}                
        \\
        & wF1              
        & bAcc             
        & wF1              
        & bAcc             
        & wF1               
        & bAcc             
        & wF1                   
        & bAcc                
        \\ 
        \hline 
        \hline
        ResNet 50                
        & 58.8            
        & 52.6            
        & 65.6             
        & 58.7            
        & 81.3             
        & 75.4   
        & 81.9\tiny{$\pm$9.7}
        & 81.5\tiny{$\pm$9.6} 
        \\
        ResNet 50 trunc          
        & 68.5            
        & 62.4            
        & 74.0             
        & 67.8            
        & 87.5             
        & 81.6   
        & 41.5\tiny{$\pm$11.7}
        & 45.9\tiny{$\pm$10.0}         
        \\
        DINOv2 ViT-S            
        & 72.9            
        & 65.6            
        & 78.5             
        & 70.8            
        & 87.7             
        & 82.0  
        & 52.5\tiny{$\pm$11.8}  
        & 54.5\tiny{$\pm$10.0}          
        \\
        DINOv2 ViT-B             
        & 71.9            
        & 64.8            
        & 77.3             
        & 69.9            
        & 87.8             
        & 81.8                      
        & 49.6\tiny{$\pm$17.3}  
        & 52.2\tiny{$\pm$14.3}
        \\
        DINOv2 ViT-L             
        & 72.2            
        & 64.9            
        & 77.8             
        & 72.4            
        & 89.1             
        & 83.5            
        & 51.5\tiny{$\pm$16.6}                      
        & 53.6\tiny{$\pm$13.5}                   
        \\
        DINOv2 ViT-G             
        & 77.8            
        & 70.4                 
        & 81.9                  
        & 74.2                 
        & 90.1                
        & 84.5                  
        & 21.1\tiny{$\pm$13.5}                       
        & 28.6\tiny{$\pm$10.2}                        
        \\
        CTransPath               
        & 80.8            
        & 73.9            
        & 83.1             
        & 76.8            
        & 88.0             
        & 82.5  
        & 60.2\tiny{$\pm$12.6} 
        & 60.9\tiny{$\pm$11.8}               
        \\
        Phikon ViT-B             
        & 83.3            
        & 76.5            
        & 85.1             
        & 78.7            
        & 88.2             
        & 82.7 
        & 81.8\tiny{$\pm$8.3}
        & 81.5\tiny{$\pm$8.5}
        \\ \hline
        DinoBloom-S  (ours)
        & 86.4            
        & 80.5            
        & 90.0             
        & 84.5            
        & 90.1             
        & 84.5            
        & \underline{93.0}\tiny{$\pm$3.0}
        & \underline{92.3}\tiny{$\pm$3.4} 
        \\
        DinoBloom-B  (ours)
        & 87.4   
        & 81.9 
        & 90.5    
        & 85.4   
        & 90.7    
        & 85.5         
        & 92.7\tiny{$\pm$2.9} 
        & 91.9\tiny{$\pm$3.1}
        \\
        DinoBloom-L  (ours)                      
        & \underline{88.9}				         
        & \underline{83.2}
        & \underline{91.3}
        & \underline{86.1}
        & \underline{91.2}
        & \underline{86.0}
        & 91.7\tiny{$\pm$2.4}
        & 91.0\tiny{$\pm$2.7}
        \\
        DinoBloom-G  (ours) 
        & \textbf{89.1}   
        & \textbf{83.5}  
        & \textbf{91.4}            
        & \textbf{86.4}           
        & \textbf{91.8}            
        & \textbf{86.6}           
        & \textbf{93.1}\tiny{$\pm$2.5}         
        & \textbf{92.4}\tiny{$\pm$2.8}             
        \\ \hline
    \end{tabular}
\end{table}

\textbf{Downstream evaluations.} 
In all downstream experiments, we compare the following feature extractors: 
(i) the non-medical-domain models ImageNet-pretrained ResNet50~\cite{resnet} (full and truncated) and (ii) the pretrained DINOv2 checkpoints, trained on LVD-142M~\cite{oquab2023dinov2};
(iii) the medical-image domain feature extractors CTransPath~\cite{ctranspath}, trained on 14M patches from TCGA and PAIP, 
(iv) the Phikon ViT-B model~\cite{phikon}, trained on PanCancer40M from TCGA;
and (v) our models DinoBloom-S, -B, -L, and -G. 

We evaluate the performance of all supervised classifier models by linear probe and \textit{k}-nearest neighbors (\textit{k}-NN) for cell-type classification and in a weakly-supervised multiple instance learning (MIL) setting for AML subtyping. For linear probe, the sklearn LogisticRegression class is used with l2-regularization coefficient of $\frac{c \times n}{100}$ where $n$ is the number of training samples, and $c$ is the number of classes.
For the MIL evaluation, similar to the Hehr et. al~\cite{hehr2023explainable} framework, we deploy a dedicated classifier head, structured as a two linear layer architecture with an intermediary ReLU activation, tailored to map the aggregated latent vectors of a patient to a class prediction.
\\

%% file: preprint/results.tex
\begin{table}[t]
    \centering
    \caption{Evaluation on bone marrow: WBC classification on the dataset BMC with 21 highly imbalanced classes.}\label{tab:bmcyto}
    \begin{tabular}{l|ccc|ccc|ccc}
        \hline
        & \multicolumn{3}{c|}{1-NN}              
        & \multicolumn{3}{c|}{20-NN}            
        & \multicolumn{3}{c}{Linear probe}              
        \\
        & wF1   
        & Acc            
        & bAcc  
        & wF1   
        & Acc            
        & bAcc  
        & wF1   
        & Acc            
        & bAcc  
        \\ \hline \hline
        ResNet 50                    
        & 37.6          
        & 37.3          
        & 21.1          
        & 47.4          
        & 50.0          
        & 23.0          
        & 64.1          
        & 65.2          
        & 39.6          
        \\
        ResNet 50 trunc              
        & 46.7          
        & 46.4          
        & 31.2          
        & 57.5          
        & 59.8          
        & 33.0          
        & 74.5         
        & 75.0         
        & 49.8       
        \\
        DINOv2 ViT-S     
        & 43.2        
        & 43.2        
        & 25.0         
        & 52.4        
        & 55.6         
        & 26.5         
        & 68.1         
        & 69.0          
        & 44.9         
        \\
        DINOv2 ViT-B            
        & 39.8        
        & 39.6         
        & 23.9        
        & 49.3        
        & 55.6       
        & 24.1        
        & 70.8         
        & 71.5        
        & 48.5         
        \\
        DINOv2 ViT-L        
        & 39.4             
        & 39.2      
        & 24.5             
        & 48.9            
        & 52.2            
        & 24.0
        & 71.0
        & 71.6
        & 47.8
        \\
        DINOv2 ViT-G  
        & 41.0								        
        & 41.0
        & 22.6
        & 50.4
        & 53.7
        & 24.4
        & 73.5
        & 74.0
        & 52.1 
        \\
        CTransPath     
        & 49.1       
        & 48.7        
        & 42.0        
        & 58.5         
        & 60.3       
        & 36.1        
        & 74.1        
        & 74.9        
        & 52.2         
        \\
        Phikon ViT-B       
        & 47.5         
        & 47.2         
        & 40.8         
        & 57.1         
        & 59.0         
        & 35.5          
        & 73.2         
        & 73.8         
        & 54.4        
        \\ \hline
        DinoBloom-S (ours)     
        & 78.4         
        & 78.3          
        & \underline{62.0}         
        & \textbf{84.2}
        & \textbf{84.8} 
        & 55.6          
        & \textbf{85.7} 
        & \textbf{85.9}
        & \textbf{71.4} 
        \\
        DinoBloom-B (ours)      
        & \underline{79.6}
        & \underline{79.5}
        & \textbf{65.8} 
        & 83.7
        & 84.1
        & \textbf{57.1} 
        & \underline{85.5}          
        & \underline{85.6}         
        & \underline{70.7}
        \\        
        DinoBloom-L (ours)        
        & 78.8
        & 78.8							
        & 57.7
        & 83.6
        & 84.0
        & \underline{56.3}
        & 84.9
        & 85.0
        & 64.4
        \\
        DinoBloom-G (ours)          
        & \textbf{80.0}
        & \textbf{79.9}
        & 59.4
        & \underline{83.8}
        & \underline{84.2}
        & 56.2
        & 84.9
        & 85.0
        & 69.3
        \\
        \hline
    \end{tabular}
\end{table}

\begin{figure}[t]
    \includegraphics[width=\textwidth]{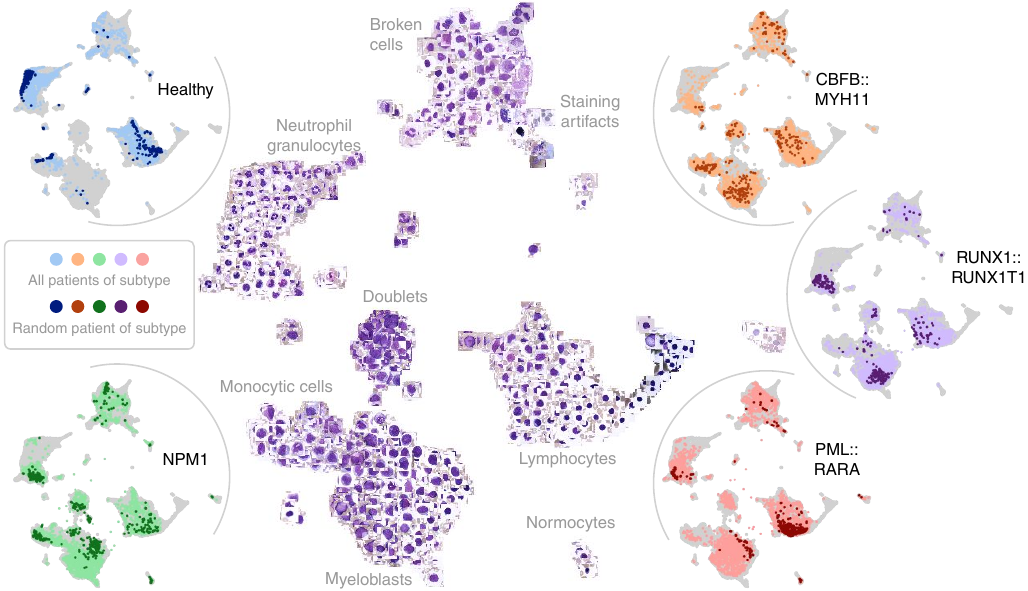}
    \caption{Low dimensional representation (UMAP) of DinoBloom-B features of over 80,000 single cells from the training set of the dataset AML Hehr. Center: UMAP with original images. Five arcs: UMAP for healthy patients (blue) and patients with CBFB::MYH11 (orange), NPM1 (green), PML::RARA (red), and RUNX1::RUNX1T1 (purple), for every class: all patients in the test set (bright) and embedding of one random test patient (dark). The myeloblast cluster and doublet cluster are barely populated for healthy controls. Different AML entities present with distinct cell patterns within the embedding. } \label{fig:umaps}
\end{figure}

\section{Results}
\textbf{Peripheral blood.}
DinoBloom models outperform existing models on single WBC classification on the external dataset Acevedo by a large margin in all variants. There is a strong performance gain over the original DINOv2 models, e.g., ViT-S with $71.9$ weighted F1-score on 1-NN vs.\ DinoBloom-S $86.4$, ViT-G $77.8$ vs.\ DinoBloom-G $89.1$  (\Cref{tab:hema}). 
The histopathology domain-specific feature extractors CTransPath and the recently released Phikon (Vit-B) model perform better than models with non-medical pretraining, both models perform roughly equally in linear probing, while Phikon has a slightly better performance in k-NN evaluations. 
However, DinoBloom models do not only perform better than their corresponding baseline from DINOv2, but even our smallest model performs better than all other tested models, irrespective of their size. One can also observe that the larger variants of DinoBloom models perform better compared to smaller versions, e.g., DinoBloom-G vs.\ DinoBloom-B vs.\ DinoBloom-S with the weighted F1-score on 1-NN (89.1 vs.\ 87.4 vs.\  86.4).

DinoBloom effectively serves as a feature extractor for training a weakly-supervised AML subtype classifier with ABMIL~\cite{ilse2018attention} aggregation.
%
%
In our experiments, DinoBloom models achieve a weighted F1-score between $91.7$ (DinoBloom-L) and $93.1$ (DinoBloom-G), while the second best models are ResNet50 and Phikon Vit-B with 81.9 and $81.8$, respectively. 

\textbf{Bonemarrow cytology.} 
In line with the results on the Acevedo dataset, DinoBloom models outperform both non-medical and medical models on bone marrow WBC classification by even larger margins.
The classification task on BMC is heavily imbalanced with $21$ classes and over $170,000$ samples in total, where some classes have a very low sample count, such as abnormal eosinophils ($8$) or smudge cells ($42$,~\Cref{fig:overview}e). 
Despite the challenging task, DinoBloom-B reaches a balanced accuracy of $65.8$ and an accuracy of $79.5$ in 1-NN evaluation compared to $42.0$ and $48.7$ of the next best model (\Cref{tab:bmcyto}). 
Similar large gaps in performance are also observed in linear probing, where our best model, DinoBLoom-S, achieves a balanced accuracy of $71.4$ compared to $52.2$ of CTransPath. Notable is also the doubling of performance in 1-NN weighted-F1 compared to the DINOv2 baseline that can be observed for the ViT-B variant. Compared to the peripheral blood tasks, the performance of the larger variants of DinoBloom is not consistently better than that of the smaller variants.

\textbf{Patient embeddings.}
We show a potential clinical application of our model: The low-dimensional embedding of AML and healthy patients from the training set of the AML Hehr dataset shows that related cell types cluster well (\Cref{fig:umaps}). 
Based on this fit, new patients (from the test set) are embedded into the same lower dimensional feature representation. The distribution of cell types is clearly distinct between healthy patients and all AML subtypes: as expected there are almost no cells in the Myeloblast related cluster in the healthy group. More subtle differences can be seen between clustering profiles of distinct AML subtypes, e.g., CBFB::MYH11 and RUNX1::RUNX1T1.
The cell embeddings of our model could give clinicians an easy-to-grasp visualization of the cell distribution of a patient and help to verify the manual quantification of different cell types. For instance, experts could identify the presence of myeloblasts within an entire smear, and gates could be applied to the embedding to facilitate morphology-based cell counting, similar to FACS analysis. 
%
%

\begin{figure}[t]
    \centering
    \includegraphics[width=\textwidth]{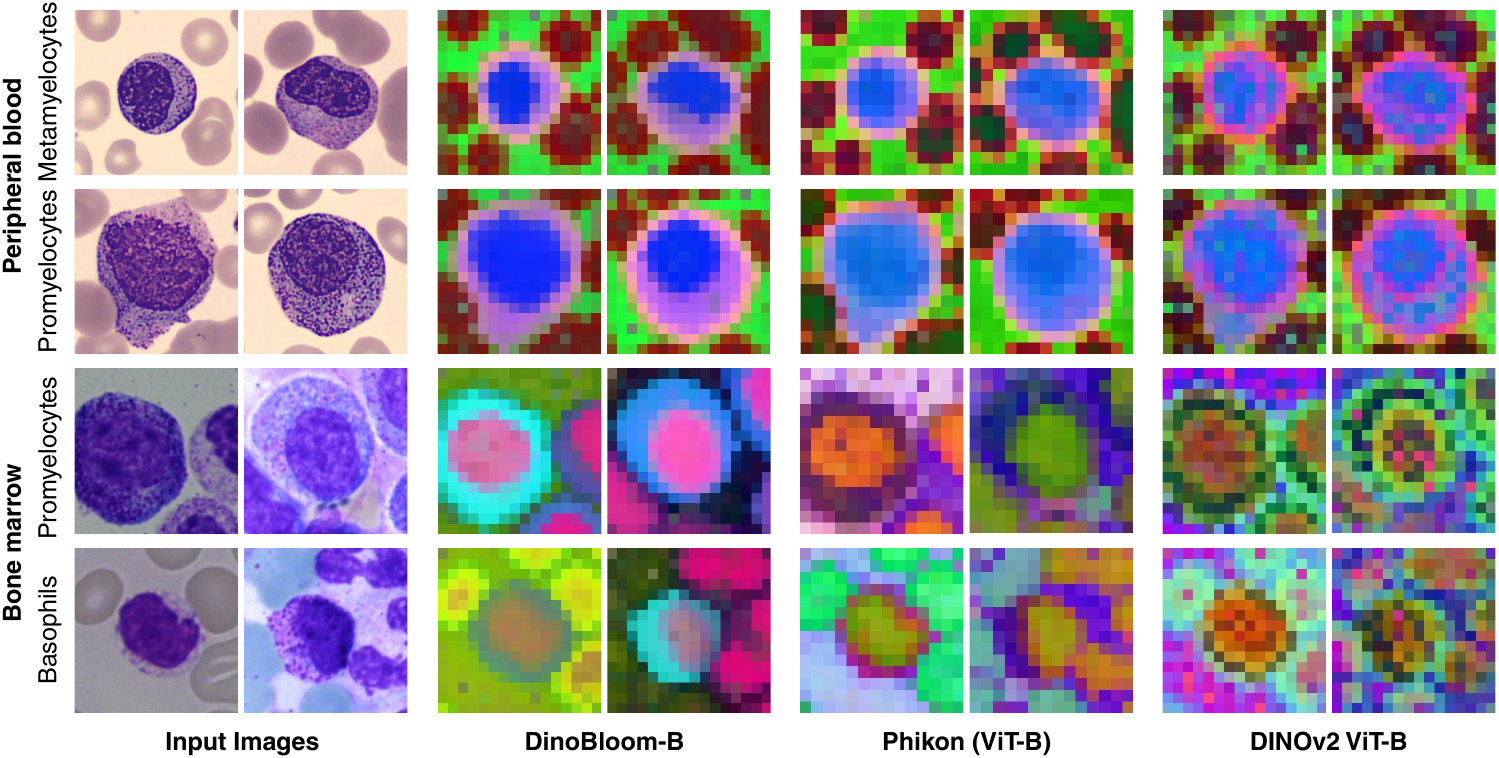}
    \caption{PCA visualization of the patch tokens on the test data of Acevedo (external) and BMC. Comparison between DinoBloom-B, the second best model Phikon (ViT-B), and the pretrained DINOv2 ViT-B. Colors represent the values of the first three PCA components. DinoBloom-B can differentiate between nuclei, cytoplasm, surrounding red blood cells, and background.}
    \label{fig:pca}
\end{figure}
\textbf{Interpretability.} In~\Cref{fig:pca}, we show that our model learns robust and meaningful features across domains and compare it to its baseline, the pretrained DINOV2, and the second best performing model Phikon. We compute the principal components for the encoded patches of four images per dataset and visualize the first three components of each patch of size 14$\times$14 (DinoBloom-B, DINOv2 Vit-B) and 16$\times$16 pixels (Phikon), respectively, in RGB colors.  One can clearly observe that DinoBloom captures shapes of nuclei, cell body, and the surrounding cells. While the other two models also capture shape and roughly outline cells, they do not catch fine grained details, as can be especially seen in the case of the nuclei, that only DinoBloom is able to differentiates from the whole cell.

%% file: preprint/conclusion.tex
\section{Conclusion}
With DinoBloom, we introduce a publicly available family of foundation models for single cell images in hematology, trained on the largest multi-cohort dataset with over 380,000 WBC images of 13 datasets. We show its strong generalization capabilities to external datasets despite strong batch effects. 
%
Our experiments demonstrate that DinoBloom extracts rich features from hematology images, with its effectiveness demonstrated for cell-type classification and AML subtyping compared to non-medical and medical vision models. 
We also support this claim through visualizations showing that our model detects important hematological concepts, such as nuclei, cytoplasm, and red blood cells, which could be further leveraged for zero-shot segmentation. We believe that the generalizable cell embedding capabilities of our DinoBloom models offer grerat potential in assisting clinicians in their tedious manual work of cell detection and classification.